\documentclass[conference]{IEEEtran}
\IEEEoverridecommandlockouts
\usepackage{cite}
\usepackage{amsmath,amssymb,amsfonts}
\usepackage{algorithmic}
\usepackage{graphicx}
\usepackage{textcomp}
\usepackage{pdfpages}
\usepackage{url}
\usepackage[breaklinks=true]{hyperref}
\usepackage{breakcites}
\usepackage{microtype}

\def\BibTeX{{\rm B\kern-.05em{\sc i\kern-.025em b}\kern-.08em
    T\kern-.1667em\lower.7ex\hbox{E}\kern-.125emX}}
\begin{document}

\title{Salt Detection Using Segmentation of Seismic Image
}

\author{\IEEEauthorblockN{
		Mrinmoy Sarkar}
\IEEEauthorblockA{\textit{Department of Electrical \& Computer Engineering} \\
\textit{North Carolina A\&T State University}\\
Greensboro, NC-27411,  USA \\
msarkar@aggies.ncat.edu}
}

\maketitle

\begin{abstract}
In this project, state-of-the-art deep convolution neural network (DCNN) is presented to segment seismic image for salt detection below the earth surface. Detection of salt location is very important for starting mining. Hence, a seismic image is used to detect the exact salt location under the earth surface. However, precisely detecting the exact location of salt deposits is difficult. Therefore, professional seismic imaging still requires expert human interpretation of salt bodies. This leads to very subjective, highly variable renderings. Hence, to create the most accurate seismic images and 3D renderings, we need a robust algorithm that automatically and accurately identifies if a surface target is salt or not. Since the performance of DCNN is well-known and well-established for object recognition in image, DCNN is a very good choice for this particular problem and being successfully applied to a dataset of seismic images in which each pixel is labeled as salt or not. The result of this algorithm is promising.
\end{abstract}


\section*{Objectives}
\begin{itemize}
	\item  Segmentation of seismic image into salt or sediment using DCNN.
	\item Automate the process of analysis of seismic image.
	\item Reduce the cost of identifying an earth surface before mining.
\end{itemize}

\section{Introduction}
A seismic image is generated from imaging the reflection coming from rock boundaries. The seismic image shows the boundaries between different rock types. In theory, the strength of reflection is directly proportional to the difference in the physical properties on either side of the interface. While seismic images show rock boundaries, they don't say much about the rock themselves; some rocks are easy to identify while some are difficult. There are several areas of the world where there are vast quantities of salt in the subsurface. One of the challenges of seismic imaging is to identify the part of subsurface which is salt. However, it is an image segmentation problem from the image processing perspective. There are many robust algorithms available for this task in the literature such as feature-space, image-domain, and  physics-based techniques \cite{lucchese2001colour}. These techniques have been successfully used for color image segmentation captured  the digital camera. Since seismic images are significantly different than digital images, those state-of-the-art techniques fail for segmentation task. There are many challenges for seismic image segmentation. Some of them are listed as follows:
\begin{itemize}
	\item Image capturing method.
	\item Uneven distribution of salt and other rocks.
	\item Rock which have density compared to salt.
	\item Only gray-level image means lack of information.
	\item Uneven structure of rocks below the earth surface which causes uneven reflection. 
\end{itemize}

However, there are many state-of-the-art machine learning technique that can be used to solve this problem. The most promising technique in the literature is deep convolution neural network for any task related to an image. This technique has been used for object recognition \cite{ren2015faster}, image segmentation \cite{chen2018deeplab}, style transformation \cite{etemad20093d}, human action recognition \cite{ji20133d}, medical image segmentation \cite{milletari2016v} and image denoising \cite{zhang2017beyond}. Hence, the method has been used to solve the problem at hand. In this work, the contributions are as follows:

\begin{itemize}
	\item used state-of-the-art DCNN to segment seismic image for salt identification.
	\item automated the post analysis of seismic images.
	\item reduced the cost for seismic image analysis.
	\item relaxed the necessity of human expert for seismic image segmentation.
\end{itemize}

The rest of the paper is organized as in section \ref{Literature survey} literature survey, in section \ref{Methodology} our method to solve the segmentation problem, in section \ref{Experimental Results} experimental results of our method and concludes with conclusion \& future work in section \ref{Conclusion Future Work}.

\begin{figure*}[!t]
	\centerline{\includegraphics[width=7in, height=3.3in]{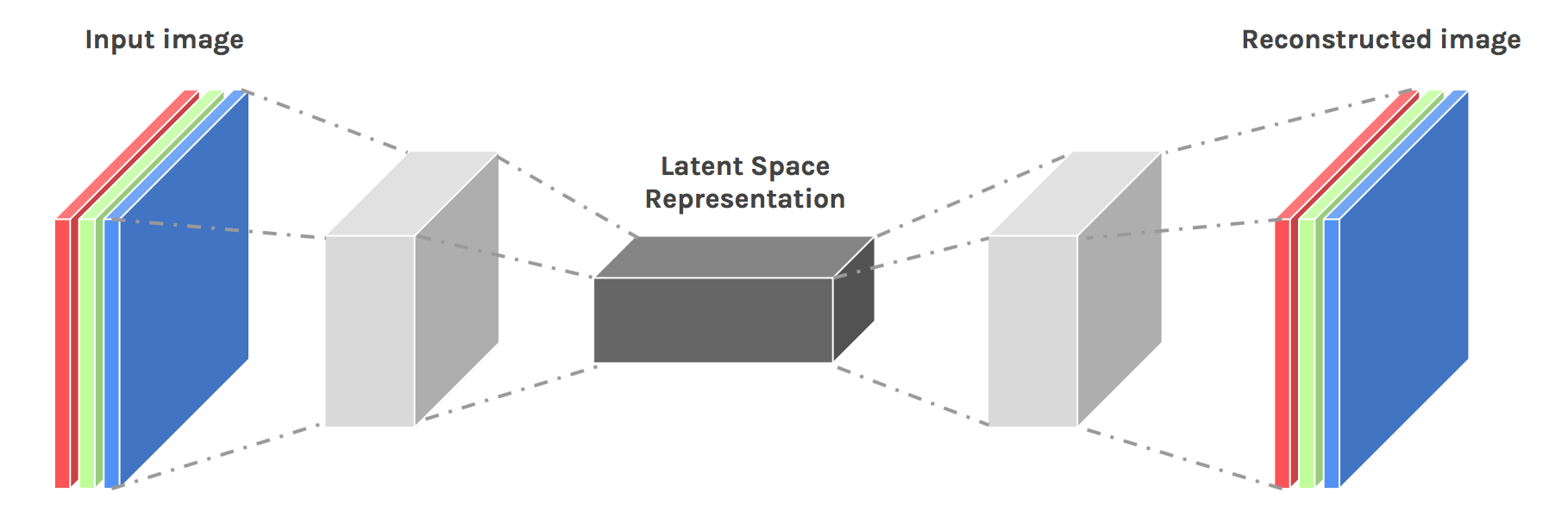}}
	\caption{Convolutional Auto-encoder architecture.\cite{autoencoder}}
	\label{fig:CAE}
\end{figure*}

\section{Literature survey}\label{Literature survey}
Image segmentation is a fundamental task for many image processing, video analysis or computer vision application. Hence, many research papers have been published on this topic. All the proposed method can be categorized into the following three techniques \cite{lucchese2001colour}.
\begin{enumerate}
	\item Feature-Space Based Techniques
	\item Image-Domain Based Techniques
	\item Physics Based Techniques
\end{enumerate}

\subsection{Feature-Space Based Techniques} 
In this approach, color is assumed to be a constant property of the surface of each object within an image. So, every pixel can be clustered or grouped into some region within the image, which will produce the segmented image. Clustering and histogram thresholding are two well-known feature-space-based techniques \cite{lucchese2001colour}.

\subsection{Image-Domain Based Techniques} 
The feature space based algorithms work on the global property of the image which satisfies the homogeneity requirement of image segmentation. However, those techniques do not consider the spatial characteristic of the image. Thus researchers found the image domain based technique. These techniques satisfy both feature-space homogeneity and spatial compactness at the same time. The spatial compactness is ensured either by subdividing and merging or by progressively growing image regions, while the homogeneity is adopted as a criterion to direct these two processes. According to the strategy preferred for spatial grouping, these algorithms are usually divided into split-and-merge and region growing techniques.
Neural-network-based classification, split and merge using region adjacency graph(RAG)  and edge-based algorithms are known to be image domain based techniques \cite{lucchese2001colour}.

\subsection{Physics Based Techniques} 
The discussed techniques are prone to segmentation error when the image is affected by highlights, shadowing and shadows. The problem can be solved considering the interaction of light with colored materials and to introduce models
of this physical interaction in the segmentation algorithms. This is the reason these techniques are known as physics-based techniques. The mathematical tools used in these techniques are quite similar to the previous two types of technique; the major difference to those is the underlying physical model developed for the reflections properties of colored matter \cite{lucchese2001colour}.

As I am using a seismic image, the images are not affected by shadowing or shadows. Our techniques lie in the image domain based techniques. However, the architecture and methodology are entirely different than the techniques discussed in \cite{lucchese2001colour}.

\section{Methodology}\label{Methodology}

In this project, I have used the auto-encoder(AE) architecture of the neural network. Auto-encoders (AE) are a family of neural networks for which the input is the same as the output. They work by compressing the input into a latent-space representation and then reconstructing the output from this representation. If the architecture is built upon convolution neural network then it is known as Convolutional Auto-encoder (CAE). Since our input is an image, I used CAE to develop my segmentation model. The CAE architecture is shown in Fig. \ref{fig:CAE}. There are two parts of this architecture, named as encoder and decoder. The encoder part consists of convolution and pooling layers and the decoder part consists of convolution and up-sampling layers. Each of these layers is describes in the following section.

\begin{figure}[!t]
	\centerline{\includegraphics[width=3in, height=2in]{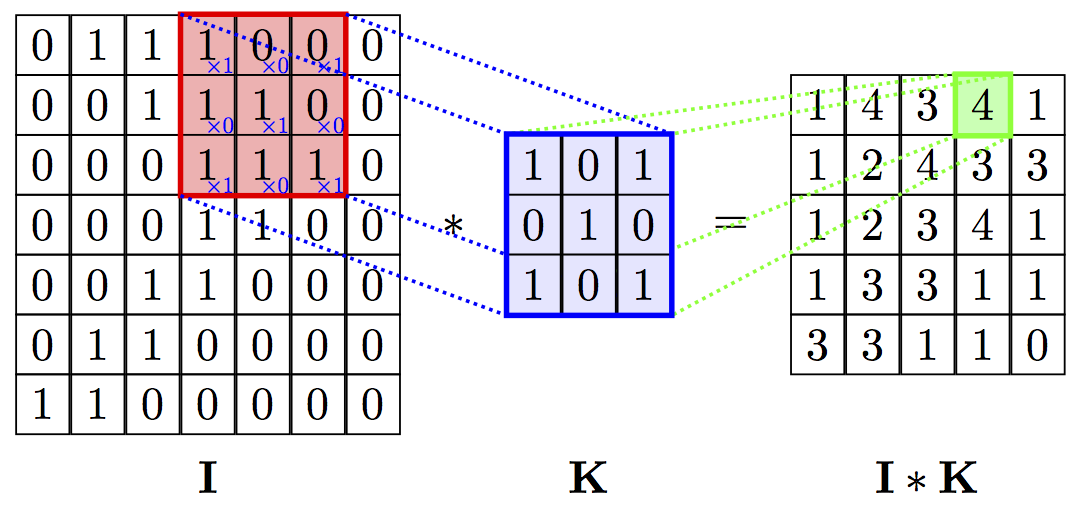}}
	\caption{Convolution operation. \cite{conv2d}}
	\label{fig:conv_layer}
\end{figure} 

\begin{figure}[!t]
	\centerline{\includegraphics[width=3in, height=2in]{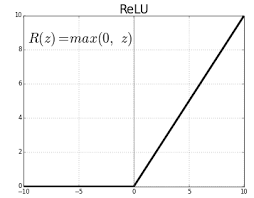}}
	\caption{ReLU activation function. \cite{relu}}
	\label{fig:relu}
\end{figure} 

\subsection{Convolution Layer} 
The main building block of any CNN architecture is the convolution layer. The convolution operation extracts the key features from the input image. This operation preserves the spatial relationship between pixels by learning image features using small squares of input data. In this layer, the input image is convolved with some predefined filters or kernel and then the output of the convolution operation is fed to an activation function. The $2D$ convolution operation is shown in equation \ref{eqn:conv}. The output of the convolution operation is known as Feature Map. The size of the Feature Map is controlled by three parameters: \cite{anintuitivecnn}
\paragraph{Depth} Depth corresponds to the number of filters used for the convolution operation.
\paragraph{Stride} Stride is the number of pixels by which the filter matrix is slid over the input matrix. 
\paragraph{Zero-padding} Sometimes zeros are padded around the border so that the filters can be applied to the bordering elements.

The convolution operation is shown in Fig. \ref{fig:conv_layer}. The last operation of a convolution layer is an activation function. The most common activation function for CNN is rectified linear unit function (ReLU). The ReLU function is shown in Fig. \ref{fig:relu}.

\begin{equation}
f(m,n)\circledast g(m,n)=\sum_{j=-\infty}^{\infty}\sum_{i=-\infty}^{\infty}f(i,j)\times g(m-i,n-j)
\label{eqn:conv}
\end{equation}

\subsection{Pooling Layer}
\begin{figure}[!t]
	\centerline{\includegraphics[width=2.5in, height=2.5in]{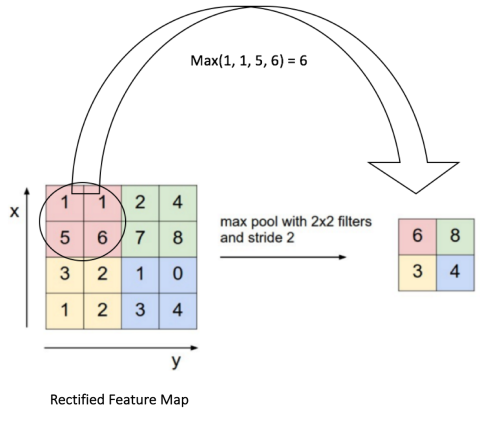}}
	\caption{Operation in pooling layer. \cite{anintuitivecnn}}
	\label{fig:pooling_layer}
\end{figure}

Another building block of a CNN is a pooling layer. It progressively reduces the spatial size of the representation to reduce the number of parameters and computation in the network. Pooling layer works on each feature map independently. Max pooling is known to be one of the most commonly used approaches. The max pooling operation is shown in Fig. \ref{fig:pooling_layer}.

\subsection{Up-sampling Layers}
\begin{figure}[!t]
	\centerline{\includegraphics[width=2.5in, height=2.5in]{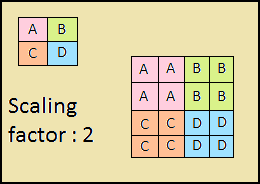}}
	\caption{Image re-sizing (Nearest-Neighbor method). \cite{nearest_neighbour}}
	\label{fig:nearest_neighbor}
\end{figure}

In this layer, the input image is re-sized to a higher dimension. There are different methods for re-sizing the image from a lower dimension to higher dimension. One of those algorithms is the nearest-neighbor algorithm. In this algorithm, the nearest pixels are copied from the original pixel value. For a scaling factor of 2, the output of the nearest neighbor algorithm is shown in Fig. \ref{fig:nearest_neighbor}.

\subsection{Architecture of the Developed Network}
The developed architecture is a combination of the three layers described above sections. The details of the architecture are given in Table \ref{tab:archi}.

\begin{table*}[!t]
	\centering
	\caption{Architecture of the developed Auto-encoder model}
	\label{tab:archi}

\begin{tabular}{lllllllll}
\hline
        &                          &                               & Encoder                                                 &                                                                                   &                                                                          &                                                                           &                                                                                  &                                                                             \\ \hline
        & \multicolumn{1}{c}{Name} & \multicolumn{1}{c}{\# Filter} & \multicolumn{1}{c}{\begin{tabular}[c]{@{}c@{}}Filter \\ dimension\end{tabular}} & \multicolumn{1}{c}{\begin{tabular}[c]{@{}c@{}}Activation\\ function\end{tabular}} & \multicolumn{1}{c}{\begin{tabular}[c]{@{}c@{}}Pool \\ size\end{tabular}} & \multicolumn{1}{c}{\begin{tabular}[c]{@{}c@{}}Stride\\ size\end{tabular}} & \multicolumn{1}{c}{\begin{tabular}[c]{@{}c@{}}Output \\ Image size\end{tabular}} & \multicolumn{1}{c}{\begin{tabular}[c]{@{}c@{}}Resize\\ method\end{tabular}} \\ \hline
layer1  & conv2d                   & 8                            & 3x3                                                                             & relu                                                                              &                                                                          &                                                                           &                                                                                  &                                                                             \\ \hline
layer2  & max\_pooling2d           &                               &                                                                                 &                                                                                   & 2x2                                                                      & 2x2                                                                       &                                                                                  &                                                                             \\ \hline
layer3  & conv2d                   & 8                            & 3x3                                                                             & relu                                                                              &                                                                          &                                                                           &                                                                                  &                                                                             \\ \hline
layer4  & max\_pooling2d           &                               &                                                                                 &                                                                                   & 2x2                                                                      & 2x2                                                                       &                                                                                  &                                                                             \\ \hline
layer5  & conv2d                   & 16                            & 3x3                                                                             & relu                                                                              &                                                                          &                                                                           &                                                                                  &                                                                             \\ \hline
layer6  & max\_pooling2d           &                               &                                                                                 &                                                                                   & 2x2                                                                      & 2x2                                                                       &                                                                                  &                                                                             \\ \hline
layer7  & conv2d                   & 16                            & 3x3                                                                             & relu                                                                              &                                                                          &                                                                           &                                                                                  &                                                                             \\ \hline
layer8  & max\_pooling2d           &                               &                                                                                 &                                                                                   & 2x2                                                                      & 2x2                                                                       &                                                                                  &                                                                             \\ \hline
layer9  & conv2d                   & 8                            & 3x3                                                                             & relu                                                                              &                                                                          &                                                                           &                                                                                  &                                                                             \\ \hline
layer10 & max\_pooling2d                  &                               &                                                                                 &                                                                                   & 2x2                                                                      & 2x2                                                                       &                                                                                  &                                                                             \\ \hline
        &                          &                               & Decoder                                                 &                                                                                   &                                                                          &                                                                           &                                                                                  &                                                                             \\ \hline
layer11 & upsampler                &                               &                                                                                 &                                                                                   &                                                                          &                                                                           & 8x8                                                                              & \begin{tabular}[c]{@{}l@{}}Nearest\\ Neighbor\end{tabular}                  \\ \hline
layer12 & conv2d                   & 8                            & 3x3                                                                             & relu                                                                              &                                                                          &                                                                           &                                                                                  &                                                                             \\ \hline
layer13 & upsampler                &                               &                                                                                 &                                                                                   &                                                                          &                                                                           & 16x16                                                                            & \begin{tabular}[c]{@{}l@{}}Nearest\\ Neighbor\end{tabular}                  \\ \hline
layer14 & conv2d                   & 16                            & 3x3                                                                             & relu                                                                              &                                                                          &                                                                           &                                                                                  &                                                                             \\ \hline
layer15 & upsampler                &                               &                                                                                 &                                                                                   &                                                                          &                                                                           & 32x32                                                                            & \begin{tabular}[c]{@{}l@{}}Nearest\\ Neighbor\end{tabular}                  \\ \hline
layer16 & conv2d                   & 16                            & 3x3                                                                             & relu                                                                              &                                                                          &                                                                           &                                                                                  &                                                                             \\ \hline
layer17 & upsampler                &                               &                                                                                 &                                                                                   &                                                                          &                                                                           & 64x64                                                                            & \begin{tabular}[c]{@{}l@{}}Nearest\\ Neighbor\end{tabular}                  \\ \hline
layer18 & conv2d                   & 8                            & 3x3                                                                             & relu                                                                              &                                                                          &                                                                           &                                                                                  &                                                                             \\ \hline
layer19 & upsampler                &                               &                                                                                 &                                                                                   &                                                                          &                                                                           & 128x128                                                                          & \begin{tabular}[c]{@{}l@{}}Nearest\\ Neighbor\end{tabular}                  \\ \hline
layer20 & conv2d                   & 8                            & 3x3                                                                             & relu                                                                              &                                                                          &                                                                           &                                                                                  &                                                                             \\ \hline
layer21 & downsampler              &                               &                                                                                 &                                                                                   &                                                                          &                                                                           & 101x101                                                                          & \begin{tabular}[c]{@{}l@{}}Nearest\\ Neighbor\end{tabular}                  \\ \hline
layer22 & conv2d                   & 1                             & 3x3                                                                             & relu                                                                              &                                                                          &                                                                           &                                                                                  &                                                                             \\ \hline
layer23 & output                   &                               &                                                                                 & sigmoid                                                                           &                                                                          &                                                                           &                                                                                  &                                                                             \\ \hline
\end{tabular}
\end{table*}

\subsection{Loss function}
To train the network, reduce-mean of sigmoid cross entropy is used as the loss function. If $x$ is the predicted label and $z$ is the true label then the sigmoid cross entropy can be written as equation \ref{eqn:sigmoid_cross_entropy} and if $z$ is a set of $n$ different values and $m$ is the total  number of training samples then the reduce mean loss can be calculated using equation \ref{eqn:loss}.

\begin{equation}
\begin{split}
&sigmoid\_cross\_entropy(x,z) \\
&= z\times (-\log(sigmoid(x)) )\\
 &+ (1-z)*(-\log(1-sigmoid(x)))
\end{split}
\label{eqn:sigmoid_cross_entropy}
\end{equation}

\begin{equation}
sigmoid(x) =\frac{1}{1+e^{-x}}
\label{eqn:sigmoid}
\end{equation}

\begin{equation}
\begin{split}
&Loss = \frac{1}{m}\times\sum_{j=1}^{m}\sum_{i=1}^{n}sigmoid\_cross\_entropy_j(x_i,z_i) 
\end{split}
\label{eqn:loss}
\end{equation} 

\subsection{Training Algorithm}
The training algorithm used for optimizing the pre-defined loss function is ADADELTA. As describes in \cite{zeiler2012adadelta}, the ADADELTA is an adaptive gradient descent algorithm which adapts the learning rate dynamically during the training process based on first order information only and the computational cost of the algorithm is less than any other state-of-the-art gradient descent algorithm. For more information about this optimization technique, readers can look into \cite{zeiler2012adadelta}.

\begin{figure}[!t]
	\centerline{\includegraphics[width=3.5in, height=2.5in]{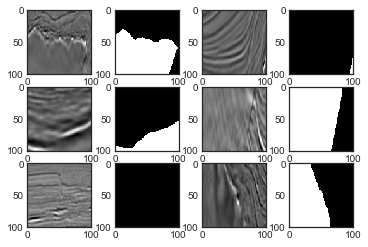}}
	\caption{Sample seismic images and corresponding mask images. }
	\label{fig:dataset}
\end{figure}
\section{Experimental Results}\label{Experimental Results}
The dataset used for this project contains 4,000 seismic images with 4,000 labeled mask images. The seismic images are in grayscale, and the mask images are in black and white. White pixel in mask image indicated the presence of salt in the original image. The size of all the images is $101\times101$. The input images are re-sized to $128\times128$ for the sake of fast computation, but the mask images are kept unchanged. Some sample seismic and mask images are shown in Fig. \ref{fig:dataset}. The data set is obtained from \cite{tsg}. Software tools used for this project, are listed below:

\begin{enumerate}
	\item Python
	\item Scikit-learn
	\item TensorFlow
	\item Keras
	\item Pandas
	\item Numpy
	\item Matplotlib
\end{enumerate}

\begin{figure}[!t]
	\centerline{\includegraphics[width=3.5in, height=2.5in]{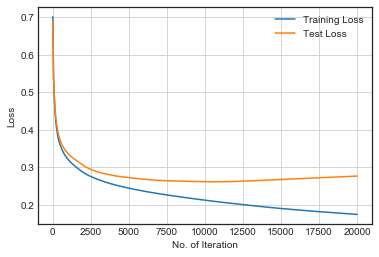}}
	\caption{Training and Testing loss vs no. of epochs. }
	\label{fig:train-test-error}
\end{figure}

I split the whole 4,000 data samples into 3,200 as training set and 800 as test set means 80\% of the data is used for training. With an initial learning rate of 0.01 and with a mini-batch size of 100, after 10,000 epochs the training loss was 0.2126, and the test loss was 0.2619. However, after 50,000 epochs the training loss was 0.1307, and the test loss is 0.1452. For one epoch, it took around 8.95 seconds. The change of training and test loss concerning epochs is shown in Fig. \ref{fig:train-test-error}.

\subsection{Cross-validation}
For validating the model, I used 10-fold cross-validation. Since the model needs a very long time for training, cross-validation is not performed from the beginning of the training. Instead, the validation is performed after 20,000 epochs, and for each fold, 500 epochs are run. The mean cross-validation error was found as 0.19.

\subsection{Prediction}
\begin{figure}[!t]
	\centerline{\includegraphics[width=3.5in, height=2.5in]{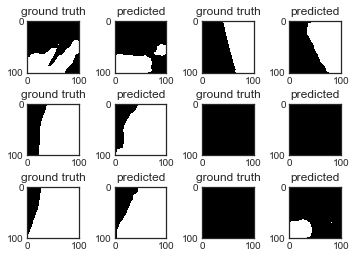}}
	\caption{Ground truth and predicted mask images. }
	\label{fig:prediction}
\end{figure}

After training the model, it is used for prediction. A threshold value of 0.5 is used to convert the probability to 0 and 1. If the output probability of any pixel is less than 0.5, it is assigned to 0 and otherwise 1. Some ground truth and predicted mask images are shown in Fig. \ref{fig:prediction}.

\section{Conclusion \& Future Work}\label{Conclusion Future Work}
In this project, I have successfully used the auto-encoder architecture of CNN to detect each pixel in a seismic image as a salt pixel or other rock. The developed technique can be used for detecting other valuable rocks embedded below the earth surface. With a proper number of epochs, the designed system can reduce the prediction error as low as 10\%. The main drawback of the proposed system is that it takes a significant amount of time to train the network. The reason is that the computational complexity of the system is very high. However, by using GPU accelerated implementation, the training time can be reduced to a reasonable time. On the other hand, the prediction operation of the trained model is very fast. Therefore, the trained model can be used for real-time detection of a scanned seismic image which will accelerate the total processing time of mining location detection. 

In the future, I will investigate how the training process can be accelerated with limited computing resources. The results of the developed technique will be compared with other existing algorithms. To conclude, I can say, the obtained result from the developed system is promising, and the trained model can be used for other seismic image analysis. The developed system is not limited to only salt detection problem.





\bibliographystyle{IEEEtran}
\bibliography{refs}

\end{document}